\documentclass[letterpaper, 10 pt, journal, twoside]{IEEEtran}

\IEEEoverridecommandlockouts                              

\usepackage{graphics} 
\usepackage{graphicx} 
\usepackage[export]{adjustbox}

\usepackage{amsmath,amssymb,amsfonts}
\usepackage{bm}  
\usepackage{caption}
\usepackage[normalem]{ulem}

\usepackage{hyperref}
\hypersetup{
    colorlinks=true,
    linkcolor=blue,
    filecolor=magenta,      
    urlcolor=cyan,
}
\usepackage{verbatim}

\title{\LARGE \bf
A Transition-Aware Method for the Simulation of Compliant Contact with Regularized Friction
}

\author{Alejandro M. Castro$^{1\,*}$, Ante Qu$^{1\,2\,*}$, Naveen Kuppuswamy$^{1}$, Alex Alspach$^{1}$, and Michael Sherman$^{1}$
\thanks{\copyright 2020 IEEE.  Personal use of this material is permitted.  Permission from IEEE must be obtained for all other uses, in any current or future media, including reprinting/republishing this material for advertising or promotional purposes, creating new collective works, for resale or redistribution to servers or lists, or reuse of any copyrighted component of this work in other works. Manuscript received: September 10, 2019; Revised December 16, 2019; Accepted January 13, 2020.}
\thanks{This paper was recommended for publication by Editor Hong Liu upon evaluation of the Associate Editor and Reviewers' comments.} 
\thanks{$^{*}$Authors have contributed equally, and names are in alphabetical
order. \hfill}%
\thanks{$^{1}$Toyota Research Institute, One Kendall Square, Binney St Building 100, Suite 1-201, Cambridge, MA 02139, USA.
   }%
\thanks{$^{2}$Stanford University, Computer Science, 353 Jane Stanford Way, Stanford, CA 94305, USA.  \newline{\tt\small[alejandro.castro, naveen.kuppuswamy,  alex.alspach, sherm]@tri.global, antequ@cs.stanford.edu}}%
\thanks{Digital Object Identifier (DOI): see top of this page.}%
}

\newcommand{\qs}{\mathbf{q}}   
\newcommand{\vs}{\mathbf{v}}   
\newcommand{\R}{\mathbb{R}}    
\newcommand{\nth}[1]{$#1\text{th}$}  

\begin{document}

\maketitle


\begin{abstract}

Multibody simulation with frictional contact has been a challenging subject of
research for the past thirty years. Rigid-body assumptions are commonly used to
approximate the physics of contact, and together with Coulomb friction, lead to
challenging-to-solve nonlinear complementarity problems (NCP). On the other
hand, robot grippers often introduce significant compliance. Compliant contact,
combined with regularized friction, can be modeled entirely with ODEs, avoiding
NCP solves. Unfortunately, regularized friction introduces high-frequency stiff
dynamics and even implicit methods struggle with these systems, especially
during slip-stick transitions. To improve the performance of implicit integration for these systems we
introduce a Transition-Aware Line Search (TALS), which greatly improves the
convergence of the Newton-Raphson iterations performed by implicit integrators.
We find that TALS works best with semi-implicit integration, but that the
explicit treatment of normal compliance can be problematic. To address this, we
develop a Transition-Aware Modified Semi-Implicit (TAMSI) integrator that has
similar computational cost to semi-implicit methods but implicitly couples
compliant contact forces, leading to a more robust method. We evaluate the
robustness, accuracy and performance of TAMSI and demonstrate our approach
alongside relevant sim-to-real manipulation tasks.
\end{abstract}

\section{Introduction}
\IEEEPARstart{R}{ecently} robotics has experienced a dramatic boom due to the introduction of new
sensor technologies, actuation, and innovative software algorithms that allow
robots to reason about the world around them. These new technologies are
allowing the next generation of robots to start moving from their highly
structured environments in factories and research labs to less structured,
richer environments such as those found in the home. There are still, however, a
large variety of research problems to be solved. In particular, manipulation is
one area of robotics that raises significant challenges including high-speed
sensing, planning, and control. 

Simulation has proven indispensable in this new era of robotics, aiding at
multiple stages during the mechanical and control design, testing, and training
of robotic systems. For instance, \cite{bib:bousmalis2018using} demonstrates the
use of simulated data to train a robotic system to grasp new objects, while
\cite{bib:chebotar2019closing} studies how to transfer policies trained in simulation to the real world.

\begin{figure}[t]
    \centering
    \adjincludegraphics[width=0.49\columnwidth,trim={{.15\width} 0 {.15\width} 0},clip]{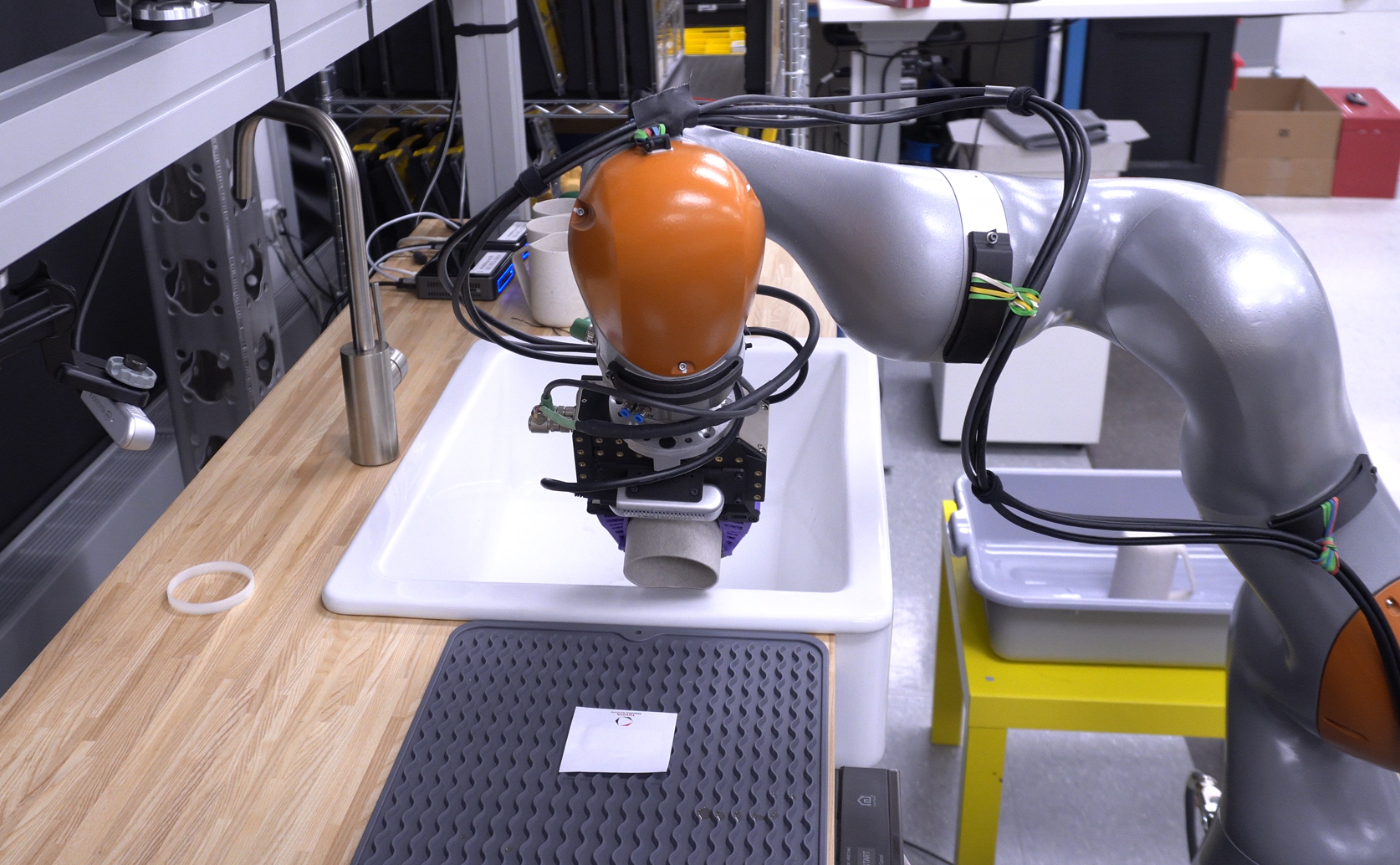}
    \adjincludegraphics[width=0.49\columnwidth,trim={{.15\width} 0 {.15\width} 0},clip]{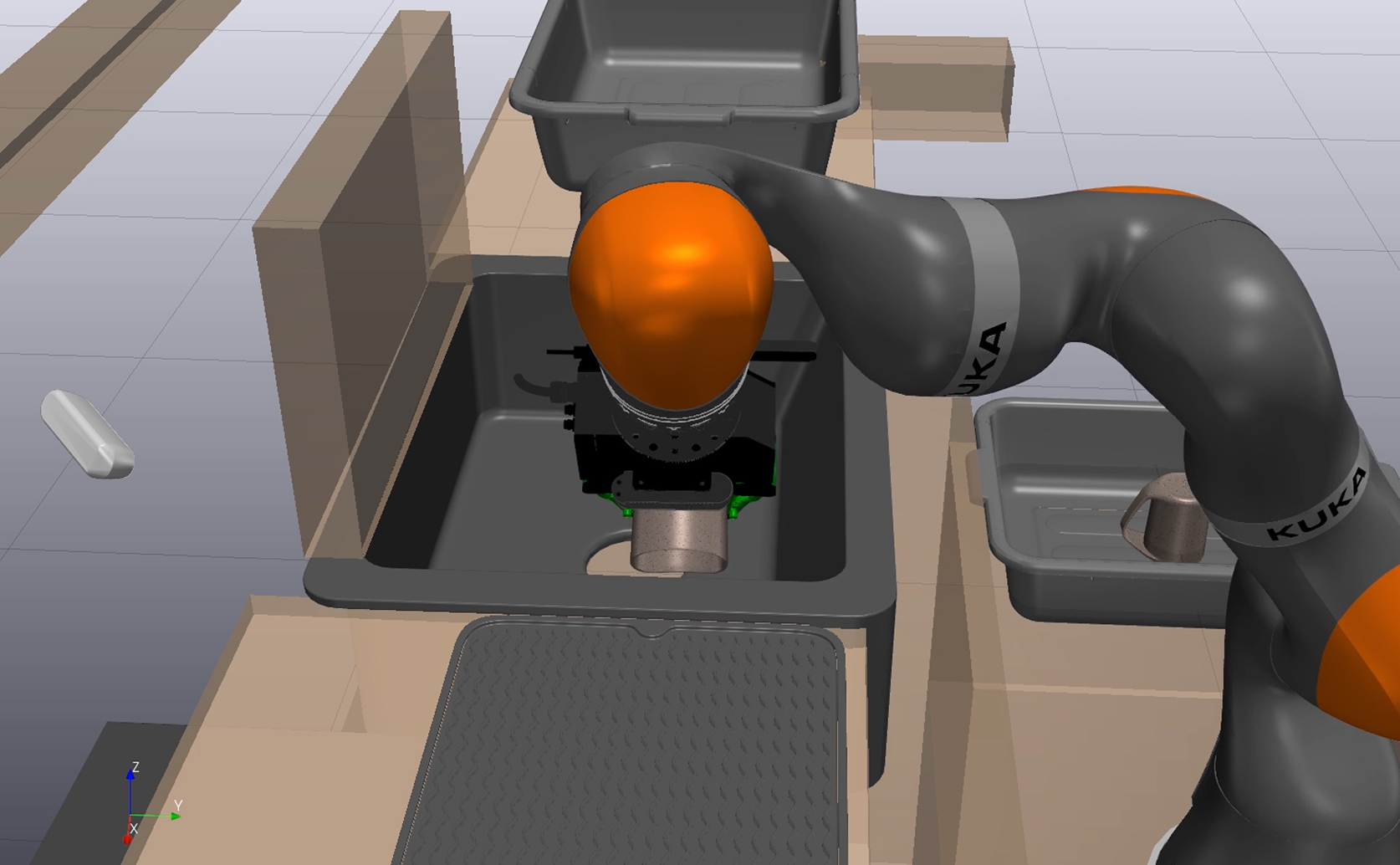}
    \caption{Sim-to-real comparison for the task of reorienting a mug by making contact with an external surface. See supplemental video.}
    \label{fig:teaser}
\end{figure}

Near real-time forward simulation has found applications in contact-aware state
estimation wherein predicted contacts and physical consistency
\cite{bib:kolev2015physically} are used to estimate the state of a robot, 
manipulands, or both \cite{bib:duff2011physical, bib:li2015comparative}.

We would like to 
synthesize, train, and validate controllers in simulation with the
expectation that they will work well in reality. Hence simulation should present
a controller with a range of \textit{physical} models and disturbances, but must
avoid significant \textit{non-physical} simulation artifacts that have no 
real-world counterparts. A central challenge in simulation for manipulation is
the physical modeling and numerical solution of multibody dynamics with contact
and friction. Such simulations
often involve high mass ratios, stiff
dynamics, complex geometries, and friction. Artifacts that are unimportant for other applications are highly amplified in the simulation of
a manipulation task; simulations either become unstable or predict highly
unstable grasps even if stable in the real system. These characteristics impose
strict requirements on robustness, accuracy, and performance to the simulation
engine when compared to other robotic scenarios with contact, such as walking.

A popular approximation to the true compliant physics of
contact is the mathematical limit in which bodies are rigid; however, it can
lead to indeterminate systems with multiple solutions, or no solution
\cite{bib:hogan2017regularization}. Still, the rigid-body approximation is at
the core of many simulation engines, enabling them to run at interactive rates.

Generally, rigid-body assumptions and Coulomb friction lead to a complex
formulation in terms of a nonlinear complementarity problem (NCP), which can be
simplified to a linear complementarity problem (LCP) using a polygonization of
the friction cone \cite{bib:stewart1996implicit, bib:anitescu1997formulating}.
LCPs are often solved using direct pivoting methods such as Lemke's algorithm.
Even though there are theoretical results for the solution of LCPs
\cite{bib:cottle1992linear}, the class of systems for which they apply are
seldom found in practice, and robust LCP implementations are either slow or
proprietary. The authors of \cite{bib:kaufman2008staggered} iteratively solve a quadratic
program (QP) for the friction impulses and a second QP for the normal impulses.
Although the results are promising, the coupled problem is non-convex and the
solution is not guaranteed to exist.

Another drawback of LCP formulations is that the linearization of the friction
cone might lead to preferential directions and cause bias in the 
solution \cite{bib:li2018implicit}---an example of a potentially-significant
non-physical simulation artifact. The computational gain from approximating the friction
cone is not clear given that the LCP introduces many auxiliary variables
and constraints, and thus \cite{bib:todorov2010implicit, bib:macklin2019non} propose
to solve the original NCP with a non-smooth Newton method.
It is common practice to relax contact constraints by introducing
\textit{regularization} \cite{bib:lacoursiere2011spook} or \textit{softening}
\cite{bib:catto2011soft, bib:bullet, bib:ode}, allowing objects to
interpenetrate due to \textit{numerical compliance}.

Given that the true physics of contact involves compliance and many of the
approaches using rigid-body assumptions introduce numerical compliance to make
the problem tractable, our work favors using compliant contact.
Models in the literature include those for point contact based on
Hertz theory \cite{bib:luo2006compliant}, volumetric models
\cite{bib:gonthier2007contact} and more sophisticated approaches modeling
contact patches \cite{bib:sherman2011simbody, bib:elandt2019pressure}.

Coulomb friction can also be regularized by replacing the strict friction-cone
constraint with a smooth function of the slip velocity and introducing a
\textit{regularization parameter}, or \textit{stiction velocity}, $v_s$ so
that objects supposedly in stiction still slide with a velocity smaller than $v_s$; for example see \cite{bib:baraff1991coping}.

By incorporating compliant contact forces and regularized friction, we can write
the system dynamics as a system of continuous ordinary differential equations
(ODEs). However, we observe that in our simulations of manipulation tasks the
stiffness of the model is dominated by the regularization of friction. In static
equilibrium, the forces due to friction balance the pull of gravity which would
otherwise accelerate the object downwards at $g \approx
9.8~\text{m}/\text{s}^2$. Using a typical $v_s$ value of $10^{-4}~\text{m}/\text{s}$, the characteristic time scale introduced by
regularized friction is about $\tau \approx v_s/g \approx 10^{-5}~\text{s}$,
and thus our error-controlled integrators must take steps as small as
$0.1~\mu\text{s}$ to resolve these highly stiff dynamics. As a result,
integrators spend most of the computational effort on resolving these
artificially introduced dynamics even when common grasping tasks
involve much larger time scales on the order of tenths of a second. With
implicit integration, stability theory says we should be able to take large time
steps once the system is in stiction, even with a very small $v_s$. This has proven difficult
in practice, however. In this work, we analyze the cause and present methods
that enable the realization of this theoretical promise in practice.

We organize our work as follows. Section \ref{sec:multibody_dynamics} introduces
the mathematical framework and notation. Section \ref{sec:implicit_integration}
introduces our novel Transition-Aware Line Search (TALS) in the context of
implicit integration. We
systematically assess the performance of a variety of integrators using
work-precision plots on a series of proposed canonical test problems in Section
\ref{sec:tals_results} and measure the improvement in robustness and performance
introduced by TALS. We show TALS performs best when \textit{freezing} the
configuration of the system and propose an original Transition-Aware Modified Semi Implicit (TAMSI) method in Section
\ref{sec:tamsi}. In section \ref{sec:results} we show that TAMSI handles transitions
robustly and outperforms our best implicit integrators for the simulation of
relevant manipulation tasks. In \ref{sec:sim_to_real} we demonstrate our method
in two sim-to-real comparisons using a Kuka arm manipulation station. Final remarks
and future research directions are presented in Section \ref{sec:conclusion}.

\section{Multibody Dynamics with Contact}
\label{sec:multibody_dynamics}
We start by stating the equations of motion and introducing our notation,
\begin{align}
    \dot{\mathbf{q}} &= \mathbf{N}(\qs) \vs,
    \label{eq:kinematic_mapping}\\
    \mathbf{M}(\qs)\dot{\mathbf{v}} &= \bm{\tau}(\qs, \vs) + \mathbf{J}^T_c(\qs)\mathbf{f}_c(\qs, \vs),
    \label{eq:momentum_balance}
\end{align}
where $\qs\in\R^{n_q}$ and $\vs\in\R^{n_v}$ are the vectors of generalized
positions and velocities respectively, $\mathbf{M}(\qs)\in\R^{n_v\times n_v}$ is
the system's mass matrix, $\bm{\tau}(\qs, \vs)$ is a vector of generalized
forces containing Coriolis and gyroscopic terms, gravity, externally applied
forces, and actuation, and $\mathbf{J}_c$ and $\mathbf{f}_c$ are contact Jacobians and forces,
defined in Section~\ref{sec:contact_jacobian}. We explicitly emphasize the
functional dependence of $\mathbf{f}_c(\qs, \vs)$ on the state vector
$\mathbf{x} = (\qs, \vs)$ given that we use a compliant contact model with
regularized friction. Finally, $\mathbf{N}(\qs)\in\R^{n_q\times n_v}$ in Eq.
(\ref{eq:momentum_balance}) is the block diagonal mapping between generalized
velocities $\vs $ and time derivative of generalized positions
$\dot{\mathbf{q}}$. Together, Eqs. (\ref{eq:kinematic_mapping}) and
(\ref{eq:momentum_balance}) describe the system's dynamics as
\begin{equation}
    \dot{\mathbf{x}}=\mathbf{f}(t, \mathbf{x}).
    \label{eq:multibody_system_dynamics}
\end{equation} 

\subsection{Kinematics of Contact}
\label{sec:contact_jacobian}

In point contact, contact between sufficiently smooth nonconforming surfaces
can be simply characterized by a pair of witness points $A_w$ and $B_w$ on
bodies $A$ and $B$, respectively such that $A_w$ is a point on the surface of
$A$ that lies the farthest from the surface of $B$ and vice versa. At the
\nth{i} contact we define the contact point $C_i$ to be midway between the
witness points. For a given configuration $\mathbf{q}$ of the system each
contact point is characterized by a penetration distance $\delta_i(\qs)$ and a
contact normal $\hat{\bm{n}}_i(\qs)$ defined to point from body $B$ into body
$A$. We denote with $\bm{v}_{c,i}$ the velocity of witness point $A_w$ relative
to $B_w$, which can be uniquely split into normal velocity $\bm{v}_{n,i} =
\bm{P}_{n,i}\bm{v}_{c,i}$ and tangential velocity
$\bm{v}_{t,i}=\bm{P}_{n,i}^\perp\bm{v}_{c,i}$ where $\bm{P}_{n,i} =
\hat{\bm{n}}_i\otimes\hat{\bm{n}}_i\succeq 0$ and $\bm{P}_{n,i}^\perp= \bm{I} -
\bm{P}_{n,i}\succeq 0$ are projection matrices in $\R^{3\times 3}$.

For a state with $n_c$ contact points we group the velocities $\bm{v}_{c,i}$ of
all contact points in a vector $\mathbf{v}_c\in\R^{3n_c}$. In Eq.
(\ref{eq:momentum_balance}) the contact Jacobian
$\mathbf{J}_c(\qs)\in\R^{3n_c\times n_v}$ maps generalized velocities to
contact point velocities as $\mathbf{v}_c =
\mathbf{J}_c \vs$. We group the \emph{scalar} separation velocities
$v_{n,i} = \hat{\bm{n}}_i\cdot\bm{v}_{c,i}$ into a vector
$\mathbf{v}_n\in\R^{n_c}$ so that $\mathbf{v}_n =
\mathbf{J}_n\vs$, where $\mathbf{J}_n =
\hat{\mathbf{N}}^T\,\mathbf{J}_c\in\R^{n_c\times n_v}$ is the normal
velocities Jacobian and $\hat{\mathbf{N}} =
\text{diag}(\{\hat{\bm{n}}_i\})\in\R^{3n_c\times n_c}$. For the tangential velocities we write
\begin{align}
    \mathbf{v}_t(\qs, \vs) &=\mathbf{J}_t(\qs)\vs,
    \label{eq:tangential_velocity_jacobian}
\end{align}
with $\mathbf{J}_t(\qs) = \mathbf{P}_n^\perp(\qs)\mathbf{J}_c(\qs) \in
\R^{3n_c\times n_v}$ the Jacobian of tangential velocities and
$\mathbf{P}_{n}^\perp(\qs) = \text{diag}(\{\bm{P}_{n,i}^\perp\})\in\R^{3n_c\times3n_c}$. The reader should notice the
difference in sizes for $\mathbf{v}_n\in\R^{n_c}$ grouping the scalar separation
velocities, and for $\mathbf{v}_t\in\R^{3n_c}$ grouping 3D tangential
velocities. This choice simplifies the exposition that follows.

\subsection{Compliant Contact with Regularized Friction}
\label{sec:point_contact_with_friction}
The normal component of the contact force is modeled following the functional
form proposed in \cite{bib:hunt1975coefficient}, which is continuous in both
penetration distance $\delta_i$ and rate $\dot{\delta}_i$
as
\begin{align}
    \pi_i &= k_i ( 1 + d_i\,\dot{\delta}_i )_+ {\delta_i}_+,
    \label{eq:compliant_normal_force}
\end{align}
where $(\cdot)_+ = \max(\cdot, 0)$ and $k_i$ and $d_i$ are stiffness and damping
parameters. Thus, the normal component of the contact force on body $A$ applied
at $C_i$ is $\bm{f}_{n,i} = \pi_i \hat{\bm{n}}_i$. These parameters can be
treated as either physical parameters computed for instance according to the
theory of Hertz contact as in \cite{bib:sherman2011simbody} or as numerical
penalty parameters as in \cite{bib:catto2011soft, bib:drake}. Since the
penetration distance $\delta_i$ is defined positive for overlapping geometries,
its time derivative relates to the separation velocity $v_{n,i}$ by
$\dot{\delta}_i=-v_{n,i}$.

We approximate the Coulomb friction force on body $A$ at $C_i$ with a linear
functional form of the tangential velocity $\bm{v}_{t,i}$, though smoother
functions can be used
\begin{align}
    \bm{f}_{t,i} &= -\tilde{\mu}_i(\Vert\bm{v}_{t,i}\Vert/v_s)\,\pi_i\,\hat{\bm{v}}_{t,i},
    \label{eq:regularized_friction_model}\\
    \tilde{\mu}_i(s) &= \begin{cases} \mu_i\,s, &0 \le s \le 1, \\
    \mu_i, & 1 < s,
    \end{cases}
    \label{eq:regularized_friction_coefficient}
\end{align}
where $\tilde{\mu}(s) \ge 0$ is the regularized friction coefficient with
$\mu_i$ the coefficient of friction and $v_s$, with units of velocity, is the
regularization parameter. We show in Section \ref{sec:one_way_coupling} that
regularized friction with positive slopes, i.e.\ $\tilde{\mu}'(s) \ge 0$ leads
to considerably more stable integration schemes.

As with velocities, we group contact forces $\bm{f}_{c,i}$ into a single vector
$\mathbf{f}_c(\qs, \vs)\in\R^{3n_c}$. We split the contact forces in their
normal and tangential components as $\mathbf{f}_c = \mathbf{f}_n + \mathbf{f}_t$
and define the generalized forces due to contact as $\bm{\tau}_n(\qs, \vs) =
\mathbf{J}^T_c(\qs)\mathbf{f}_n(\qs, \vs)$ and $\bm{\tau}_t(\qs, \vs) =
\mathbf{J}^T_c(\qs)\mathbf{f}_t(\qs, \vs)$. We note that, since $\mathbf{f}_t =
\mathbf{P}_{n}^\perp\,\mathbf{f}_t$ we can write
\begin{align}
    \bm{\tau}_t(\qs, \vs) &= 
    \mathbf{J}^T_c(\qs)\,\mathbf{f}_t(\qs, \vs),\nonumber\\
    &=\mathbf{J}^T_c(\qs)\left(\mathbf{P}_{n}^\perp(\qs)\mathbf{f}_t(\qs, \vs)\right),\nonumber\\
    &= \left(\mathbf{P}_{n}^\perp(\qs)\mathbf{J}_c(\qs)\right)^T\mathbf{f}_t(\qs, \vs),\nonumber \\
    &=\mathbf{J}_t^T(\qs)\mathbf{f}_t(\qs, \vs).
    \label{eq:tangential_torque}
\end{align}

\section{Implicit Integration with TALS}
\label{sec:implicit_integration}

We introduce our Transition-Aware Line Search (TALS) method to improve the
convergence and robustness of implicit integration methods when using large time
steps. We first make a brief overview of the implicit Euler method,
though TALS can be used with other implicit integrators. Consider a discrete
step of size $h$ from time $t^{n-1}$ to time $t^{n} = t^{n-1} + h$. Implicit
Euler approximates the time derivative in Eq.
(\ref{eq:multibody_system_dynamics}) using a first order backward
differentiation formula. The resulting system of equations is nonlinear in
$\mathbf{x}^n$ and can be solved using Newton's method,
\begin{align}
    \Delta\mathbf{x}^{k} &=
    -\left(\mathbf{A}^k\right)^{-1}\,\mathbf{r}(\mathbf{x}^{n, k});\nonumber\\
    \mathbf{x}^{n, k+1} &= \mathbf{x}^{n, k} + \Delta\mathbf{x}^{k},
    \label{eq:newton_iteration}
\end{align}
where $k$ denotes the iteration number, the residual is defined as
${\mathbf{r}(\mathbf{x})  = \mathbf{x} - \mathbf{x}^{n-1} -
h\,\mathbf{f}(\mathbf{x})}$, and $\mathbf{A}^k =
\nabla_\mathbf{x}\mathbf{r}(\mathbf{x}^k)$ is the Jacobian of the residual. We
use the integrators implemented in Drake \cite{bib:drake} which use the stopping
criterion outlined in \cite[\S IV.8]{bib:hairer2000solving} to assess the
convergence of Newton iterations. 

\subsection{Transition-Aware Line Search}
\label{sec:tals}

\begin{figure}
    \centering
    \vspace{6pt}
    \includegraphics[width=0.7\columnwidth]{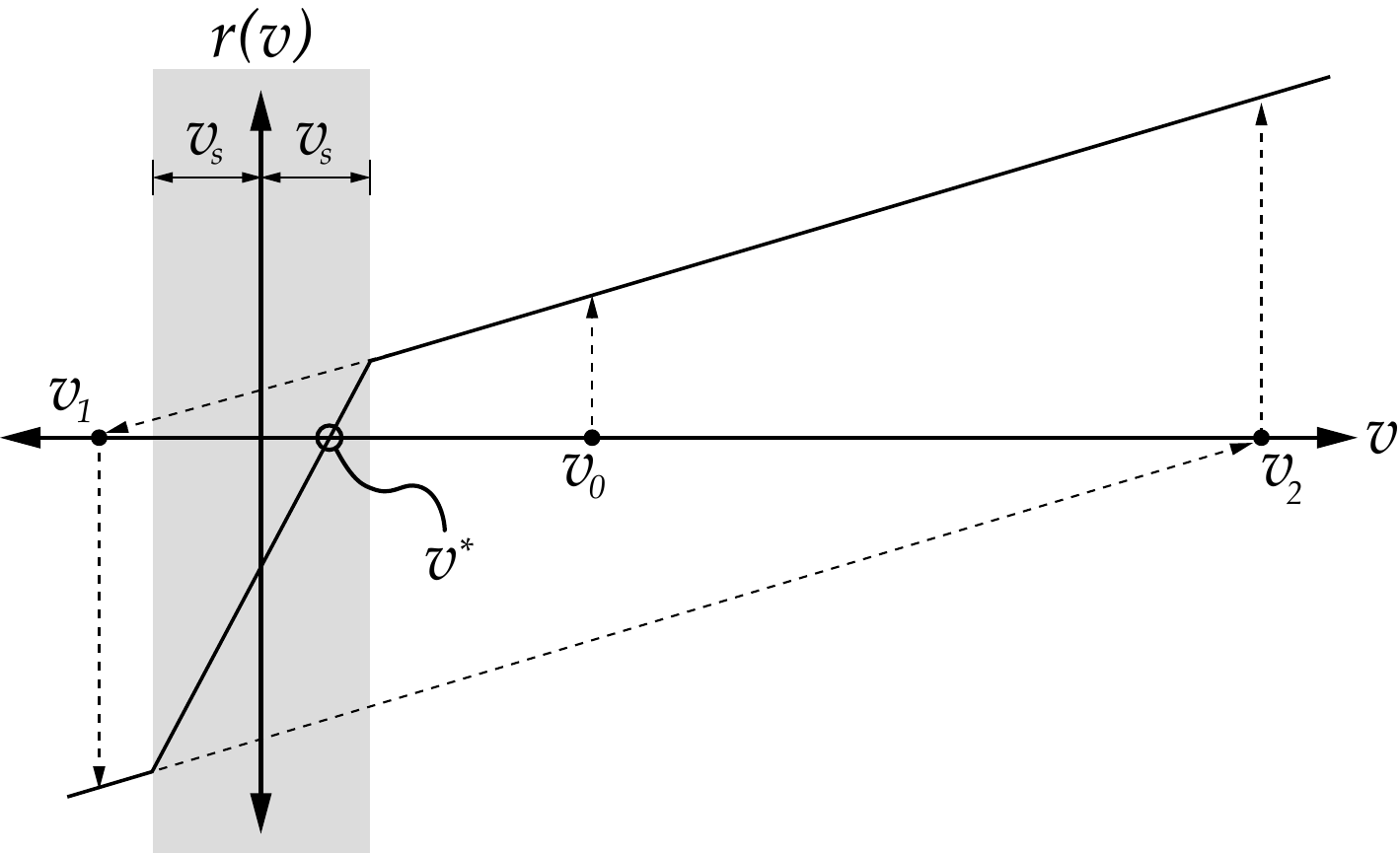}
    \caption{Divergence of Newton-Raphson near stiction. Iterations cycle
        between $v_2$ and $v_1$ indefinitely.}
    \label{fig:nrdivergence}
\end{figure}

We observed the Newton iteration in Eq. (\ref{eq:newton_iteration}) most often
fails during slip-stick transitions when using large time steps given that
$\mathbf{A}^k$ in the sliding region, $\Vert\bm{v}_{t,i}\Vert > v_s$, is not a
good approximation of $\mathbf{A}^k$ in the stiction region,
$\Vert\bm{v}_{t,i}\Vert < v_s$. This problem is illustrated in Fig.
\ref{fig:nrdivergence} for a 1D system, showing the Newton residual as a
function of sliding velocity. Notice the sharp gradient in the region
$\Vert\bm{v}_{t,i}\Vert < v_s$ due to the regularization of friction. An iterate
with positive velocity $v_0$ will follow the slope to point $v_1$ on the
negative side. Given the transition into the stiction region is missed, next
iteration then follows the slope to $v_2$ once again on the positive side.
Subsequent iterations continue to switch back and forth between positive and negative
velocities without achieving convergence. We found this can be remediated by
\textit{limiting} an iteration crossing the stiction region to fall inside of it.
As soon as $\mathbf{A}^k$ is updated with the more accurate, and larger, value
within the stiction region, Newton iterations proceed without difficulties.

\begin{figure}
    \centering
    \vspace{6pt}
    \includegraphics[width=0.45\columnwidth]{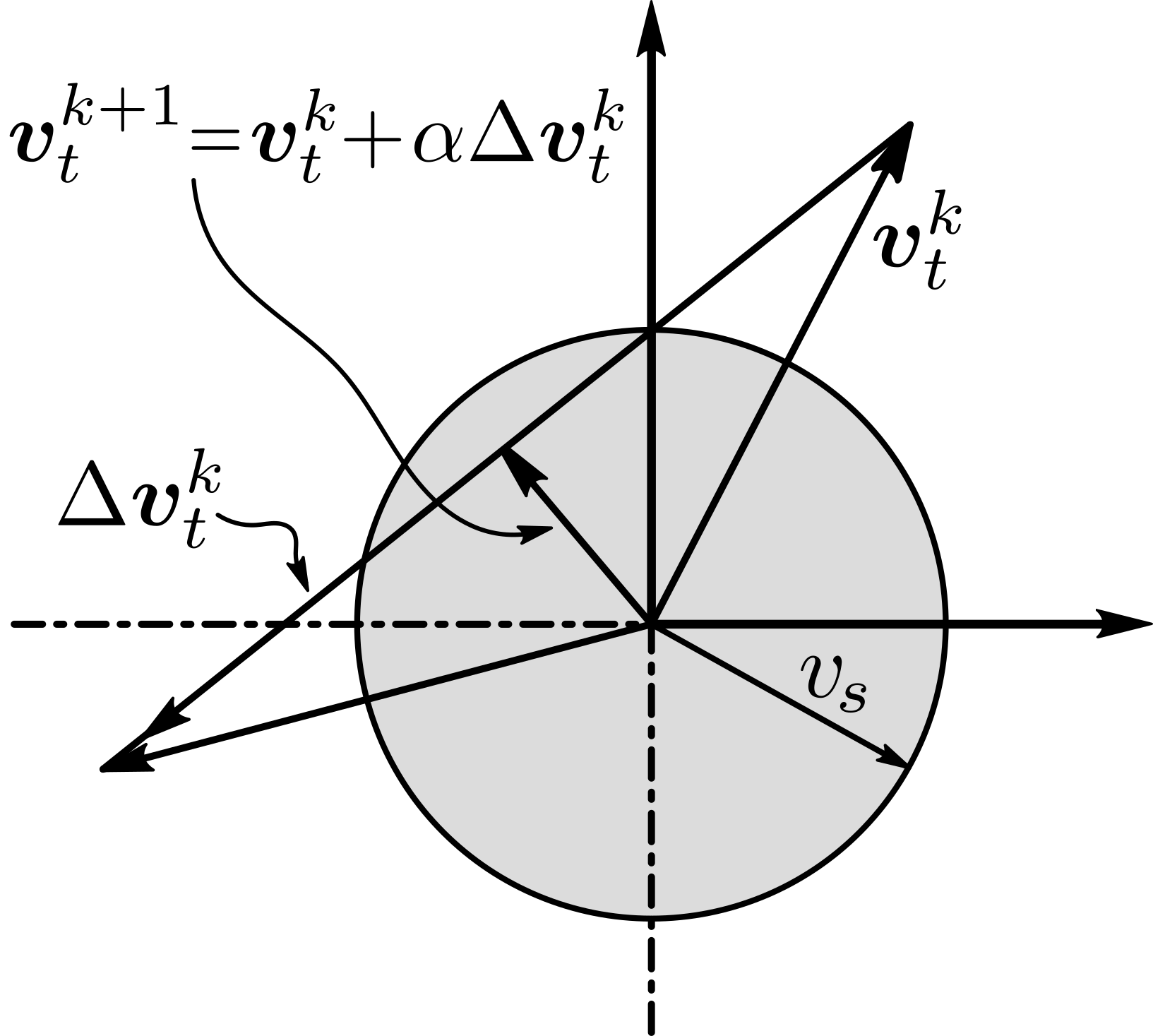}
    \includegraphics[width=0.45\columnwidth]{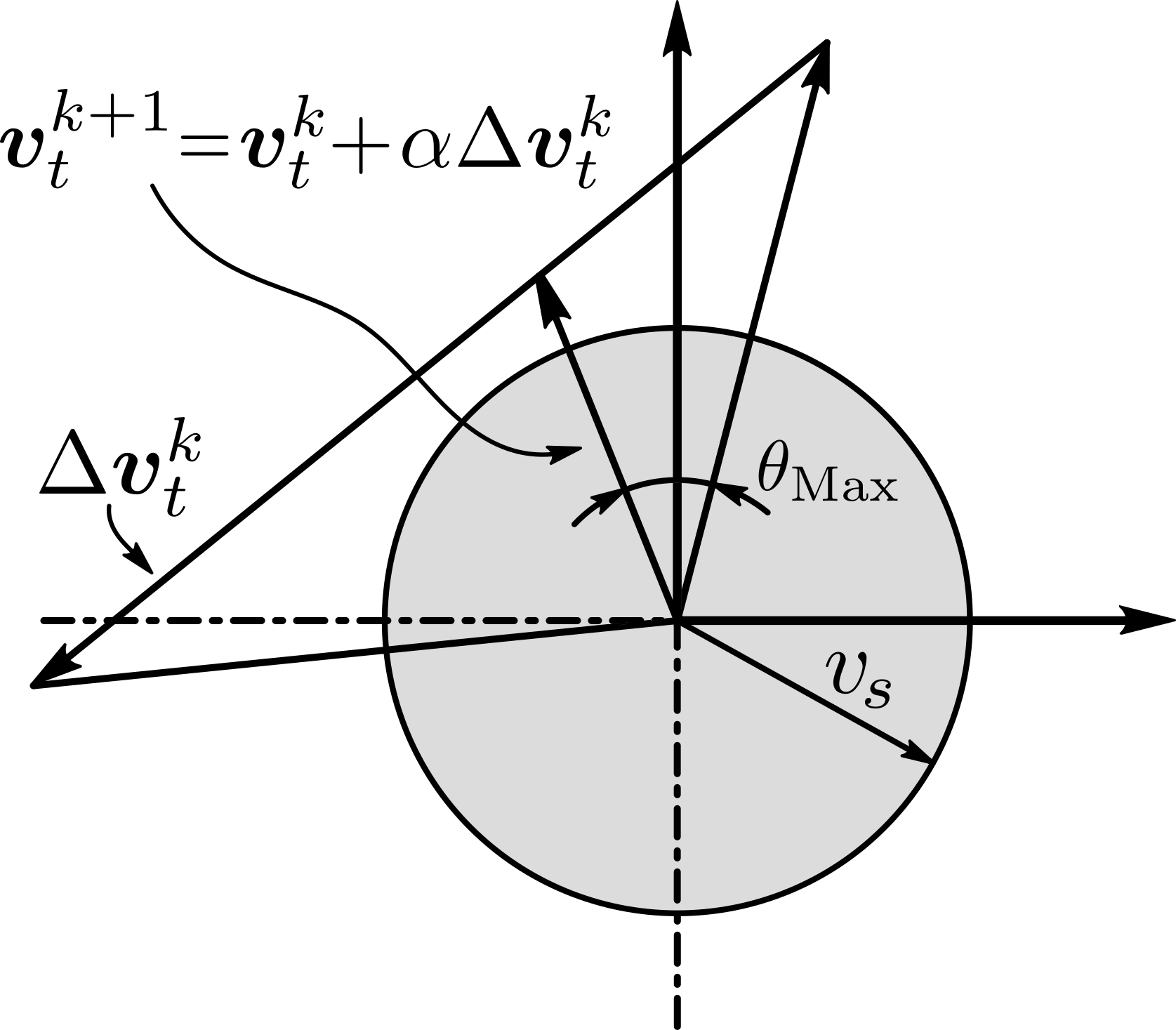}
    \caption{
        TALS limits the update of tangential velocities for the next Newton
        iteration when it detects transitions through the stiction region
        (left). We found that limiting updates to a maximum angle
        $\theta_\text{Max}$ improves robustness (right).}
    \label{fig:tals}
\end{figure}

Similar to other system-specific line search approaches
\cite{bib:todorov2010implicit,bib:zhu2018blended}, for three
dimensional problems TALS limits the iteration update in Eq.
(\ref{eq:newton_iteration}) according to $\mathbf{x}^{n,k+1} = \mathbf{x}^{n,k}
+ \alpha\Delta \mathbf{x}^{k}$ where $\alpha \in [0,1]$. If we \textit{freeze}
the configuration $\qs$ in Eq. (\ref{eq:tangential_velocity_jacobian}), this is
equivalent to limiting the tangential velocity of each contact point according to
$\bm{v}_{t,i}^{n,k+1} = \bm{v}_{t,i}^{n,k} + \alpha \Delta\bm{v}_{t,i}^{k}$, with
$\Delta\bm{v}_{t,i}^{k}$ the change predicted by Eq.
(\ref{eq:newton_iteration}), see Fig. \ref{fig:tals}. We monitor transition to
stiction by detecting the moment at which the line connecting two iterations
intersects the stiction region, inspired from the idea used in
\cite{bib:uchida2015making} to compute incremental impulses for impact problems.

TALS proceeds as follows: given a pair of iterates $\mathbf{x}^{n,k}$ and
$\mathbf{x}^{n,k+1}$ from Eq. (\ref{eq:newton_iteration}), we update tangential
velocities according to Eq. (\ref{eq:tangential_velocity_jacobian}). At the \nth{i}
contact point, if the line connecting $\bm{v}_{t,i}^{n,k}$ and
$\bm{v}_{t,i}^{n,k+1}$ crosses the stiction region, we compute $\alpha_i$ so
that $\bm{v}_{t,i}^{n,k+1} \cdot \Delta\bm{v}_{t,i}^{k} = 0$. Even if transition
is not detected, we also limit large angular changes between
$\bm{v}_{t,i}^{n,k}$ and $\bm{v}_{t,i}^{n,k+1}$ to a maximum value
$\theta_\text{Max}$, see Fig. \ref{fig:tals}. In practice we found the angular
limit to increase robustness and use $\theta_\text{Max}=\pi/3$. We finally
compute the global TALS limiting parameter as $\alpha = \min(\{\alpha_i\})$.

\subsection{Implementation Details}

We incorporate TALS in the implicit integrators implemented in Drake
\cite{bib:drake}. Drake offers \textit{error-controlled} integration to a
user-specified accuracy $a$, which translates roughly to the desired number of
significant digits in the results. We can also control whether to use a full- or
quasi-Newton method with Jacobian update strategies as outlined in \cite[\S
IV.8]{bib:hairer2000solving}. 

We make the distinction
between \textit{error-controlled} and \textit{convergence-controlled}
integration, which retries smaller time steps when Newton iterations fail to
converge, as described in \cite[\S IV.8]{bib:hairer2000solving}. All of our
fixed-step integrators use convergence control.

\section{Integration Performance}
\label{sec:tals_results}

We evaluate TALS with fixed-step and error-controlled implicit
Euler (IE) integrators. The precision obtained with fixed-step integrators is
controlled via the time step size $h$. In contrast, error-controlled integrators
adjust the time step size to meet a user-specified accuracy $a$, with additional
overhead for local error estimation.

Our objective is to evaluate the trade-off between performance and precision for
a variety of integration methods and understand if it is worth paying the additional
cost of error-controlled integration. We accomplish this by creating so-called
\textit{work-precision} plots that measure cost vs. precision in the solution.
We use the number of evaluations of the system's dynamics $\mathbf{f}(t,
\mathbf{x})$ as a metric of \textit{work}, including the evaluations used to
approximate the Jacobian of $\mathbf{f}(t, \mathbf{x})$ through forward
differencing. For \textit{precision} we want a metric that measures global error
while avoiding undesirable drifts observed over large periods, especially when
using first order methods. Therefore we \textit{localize} our global error
metric by quantifying it within a time window $\Delta_w$. To be more precise, we
introduce the \textit{flow map} $\mathbf{y} = \Phi_\tau(\mathbf{x}_0)$ such that
$\mathbf{y} = \int_0^\tau \mathbf{f} \mathrm{d}t$ with initial condition
$\mathbf{y}(0) = \mathbf{x}_0$. We compute a solution $\mathbf{x}^m$ with the
method under test at discrete times $t^m$ spaced by intervals of duration
$\Delta_w$. We then compute a \textit{localized} reference solution defined by
$\mathbf{x}_r^m = \Phi_{\Delta_w}(\mathbf{x}^{m-1})$, where notice we use the test
solution $\mathbf{x}^{m-1}$ as the initial condition of the reference solution
$\mathbf{x}_r^m$ for the next time window $\Delta_w$. In practice we integrate
$\mathbf{x}_r^m$ numerically with a much higher precision than the errors in
the solution we want to measure. Finally, we make an error estimate by comparing
$\mathbf{x}$ with the reference $\mathbf{x}_r$, see below.

\subsection{Performance Results with a Small System}

We choose a simple 2D box system that exhibits periodic stick-slip transitions.
The box of mass $m=0.33~\text{kg}$ lies on top of a horizontal surface with
friction $\mu=1.0$ and is forced to move sideways by an external harmonic force
$f(t)$ of amplitude $4~\text{N}$ and a frequency of $1~\text{Hz}$. The system's
dynamics for this case reduces to $m\dot{v} = f(t) -
\tilde{\mu}(v)W\text{sgn}(v)$, with $W$ the weight of the box in Earth's gravity
of $g=9.8~\text{m}/\text{s}^2$ and $\tilde{\mu}(v)$ as defined in Eq.
(\ref{eq:regularized_friction_coefficient}) with $v_s=10^{-4}~\text{m}/\text{s}$.

\begin{figure}
    \centering
    \adjincludegraphics[width=0.75\columnwidth,trim={0 {.09\width} 0 0},clip]{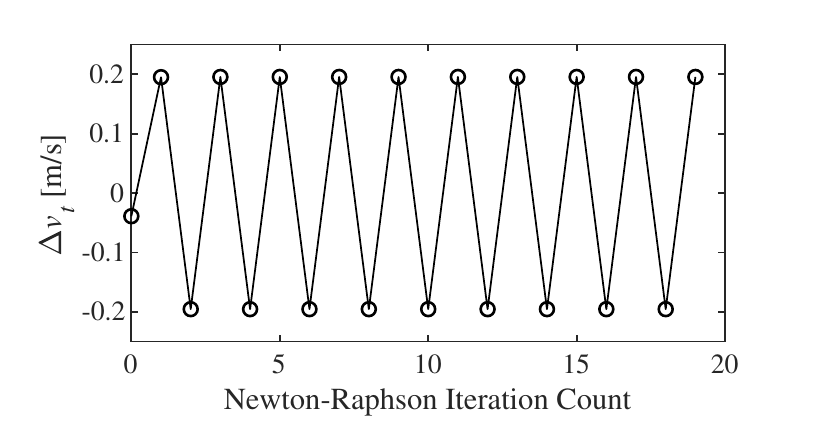}\\
    \adjincludegraphics[width=0.75\columnwidth,trim={0 0 0 0},clip]{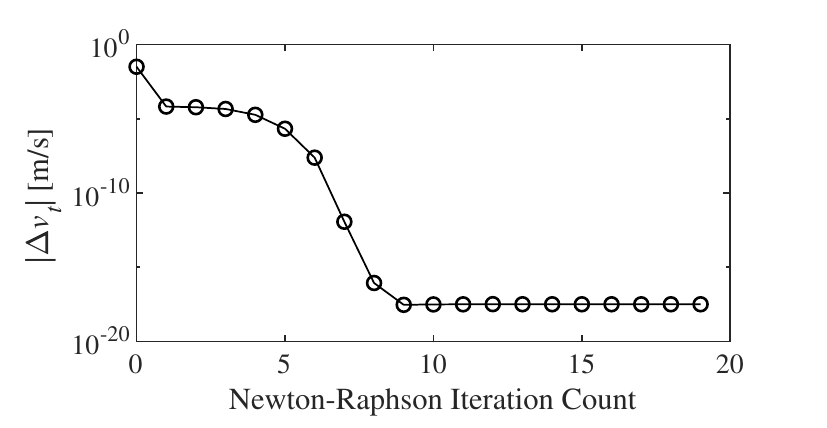}
    \caption{
        Newton's method misses stiction, oscillates, and does not converge
        (top). TALS detects the transition to stiction, limits the iteration
        update, and guides Newton's method to convergence (bottom).}
    \label{fig:nr_iterations_simple_gripper}
\end{figure}

Using a fixed time step $h=10~\text{ms}$, much larger than the time scale
introduced by regularized friction, we observe Newton iterations not to
converge as described in Section \ref{sec:tals}. Figure
\ref{fig:nr_iterations_simple_gripper} shows the Newton residual history the
first time, at $t = 0.160~\text{s}$, the box transitions from sliding into
stiction. TALS is able to detect transition, properly limit the iteration
update and even recover the second order convergence rate of Newton's method.

\begin{figure}
    \centering
    \includegraphics[scale=0.9]{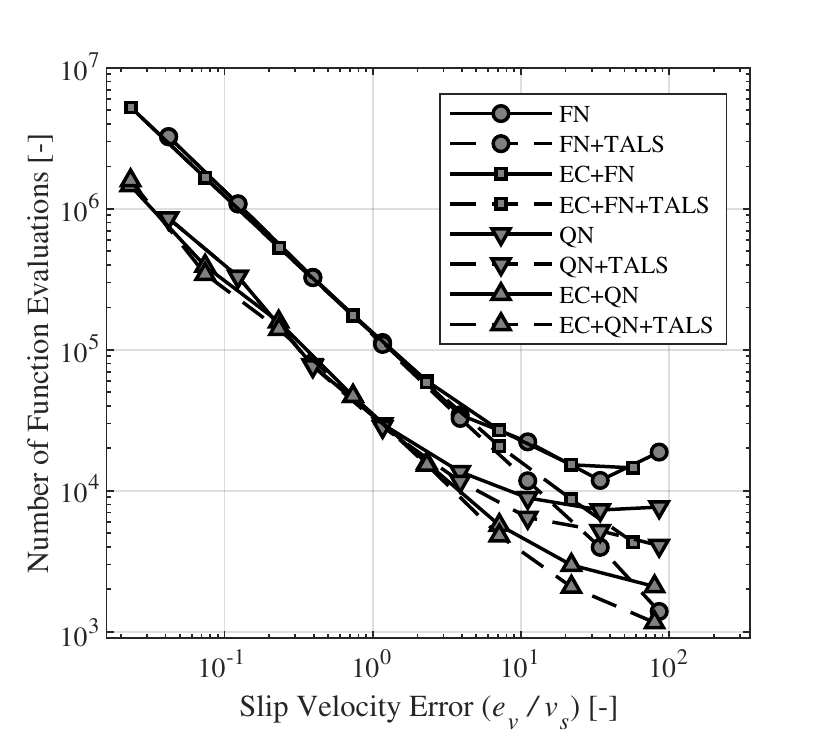}
    \caption{
        We evaluate implicit Euler integration performance with full- (FN) and quasi- (QN)
        Newton updates, each with and without error control (EC), and with and
        without TALS.
        }
    \label{fig:workprecisionimplicit}
\end{figure}

Figure \ref{fig:workprecisionimplicit} shows a work-precision plot with an error
estimate $e_v$ in the horizontal velocity $v$ defined as the $L^2\text{-norm}$
of the difference between the solution $\{v^m\}$ and the reference solution
$\{v_r^m\}$. We use a localized global error estimate computed with
$\Delta_w=25~\text{ms}$ and a reference solution computed by a $3^\text{rd}$
order Runge-Kutta with fixed step of $h=10^{-7}~\text{s}$. Each point in this
figure corresponds to a different accuracy $a$ when using error control, or a
different time step $h$ when using fixed steps. The same error metric in the
horizontal axis allows for a fair comparison.

We observe the expected theoretical first
order slope for errors near $v_s$ or smaller. However, for errors
larger than $v_s$, integrators without TALS depart from the theoretical result
as their performance degrades. Figure \ref{fig:workprecisionimplicit} shows that
TALS improves performance by extending the range in which the first order
behavior is valid, up to a factor of 10 (FN) or 3 (QN).

\subsection{Performance with Larger Systems}
\label{sec:configsensitivity}
When we applied TALS with implicit Euler to systems with more degrees of
freedom, we found TALS to be highly sensitive to changes in the configuration
$\qs$ of the system. Specifically, we observed iterations during which tangential
velocities fall outside the region $\Vert\bm{v}_t\Vert < v_s$ even after TALS
detects a transition and limits the Newton update. We traced this problem to
changes in the tangential velocities Jacobian $\mathbf{J}_t(\qs)$ in Eq.
(\ref{eq:tangential_velocity_jacobian}) caused by small changes to the
configuration $\qs$. These iterations during transition steps often result in
the same convergence failure as experienced by integrators without TALS
(Figure~\ref{fig:nr_iterations_simple_gripper}), leading to a performance
deterioration.

\section{Transition-Aware Semi-Implicit Method}
\label{sec:tamsi}

We draw from the lessons learned in the previous section to design a scheme
customized for the solution of multibody dynamics with compliant contact and
regularized Coulomb friction. Based on the observation that TALS is sensitive to
changes in the system configuration $\qs$, we decided to \textit{freeze} the
configurations in Eq. \ref{eq:momentum_balance} so that we could iterate on the
generalized velocities without changes in the configuration affecting the
stability of TALS. An important computational advantage of this approach is that
the typically demanding geometric queries only need to be performed once at the
beginning of the time step.

This is the same idea behind the semi-implicit Euler scheme. This scheme, however,
becomes unstable for stiff contact forces in the normal direction since the
position dependent terms are treated explicitly as in the conditionally stable
explicit Euler scheme. Our TAMSI scheme deals with this problem by introducing
an implicit first order approximation of the penetration distances with
generalized velocities.

We introduce TAMSI in stages to analyze the properties of
different contributions separately. We start with the traditional semi-implicit
Euler and highlight the differences as we progressively introduce TAMSI.

\subsection{Semi-Implicit Euler: One-Way Coupled TAMSI Scheme}
\label{sec:one_way_coupling}

The semi-implicit Euler scheme applied to Eqs.
(\ref{eq:kinematic_mapping})-(\ref{eq:momentum_balance}) effectively
\textit{freezes} the normal contact forces to $\mathbf{f}_{n,0}$. In this regard
the scheme is one-way coupled, meaning that normal forces couple in the
computation of friction forces but not the other way around.

Using a time step of length $h$, the semi-implicit Euler scheme applied to Eqs.
(\ref{eq:kinematic_mapping})-(\ref{eq:momentum_balance}) reads,
\begin{align}
    \qs &= \qs_0 + \mathbf{N}_0 \vs ,
    \label{eq:kinematic_mapping_discrete}\\
    \mathbf{M}_0\vs &= \mathbf{p}^* + h\mathbf{J}^T_{c,0}\mathbf{f}_c(\qs_0, \vs) ,
    \label{eq:momentum_balance_discrete}
\end{align}
where the naught subscript in $\qs_0$, $\vs_0$, $\mathbf{M}_0 =
\mathbf{M}(\qs_0)$, $\mathbf{J}^T_{c,0} = \mathbf{J}^T_{c}(\qs_0)$, and
$\mathbf{N}_0 = \mathbf{N}(\qs_0)$ denotes quantities evaluated at the previous
time step. To simplify notation we use bare $\qs$ and $\vs$ to denote the state
at the next time step. We defined $\mathbf{p}^* = \mathbf{M}_0\vs_0 +
h\bm{\tau}(\qs_0, \vs_0)$ as the momentum the system would have on the next time
step if contact forces were zero. Since we are interested in low-speed
applications to robotic manipulation, the gyroscopic terms in $\mathbf{p}^*$ are
treated explicitly. It is known however that for highly dynamics simulations,
more robust approaches should treat these terms implicitly
\cite{bib:lacoursiere2006stabilizing}.

A semi-implicit Euler scheme solves Eq. (\ref{eq:momentum_balance_discrete}) for
$\vs$ at the next time step and uses it to advance the configuration vector
$\qs$ to the next time step with Eq. (\ref{eq:kinematic_mapping_discrete}).
However this approach often becomes unstable because the stiff compliant normal
forces are explicit in the configuration $\qs$. 

To solve Eq. (\ref{eq:momentum_balance_discrete}) we define the residual 
\begin{equation}
    \mathbf{r}(\vs) = \mathbf{M}_0\vs - \mathbf{p}^* - h\bm{\tau}_{n,0} - h\bm{\tau}_{t}(\qs_0, \vs),
\end{equation}
and use its Jacobian in Newton iterations. The Jacobian is
\begin{equation}
    \nabla_\mathbf{v}\mathbf{r}(\vs) = \mathbf{M}_0 - h\nabla_\mathbf{v}\bm{\tau}_{t}(\qs_0, \vs).
    \label{eq:one_way_coupling_jacobian}
\end{equation}
Substituting in Eq.~\eqref{eq:tangential_torque},
\begin{align}
    \nabla_\mathbf{v}\bm{\tau}_{t}(\qs_0, \vs) &= \mathbf{J}^T_{t,0}\nabla_\mathbf{v}\mathbf{f}_t(\qs_0, \vs),\\
    &= \mathbf{J}^T_{t,0}\nabla_{\mathbf{v}_t}\mathbf{f}_t(\qs_0, \vs)\mathbf{J}_{t,0}, \label{eq:jacobian_of_tangential_torque}
\end{align}
where we used Eq. (\ref{eq:tangential_velocity_jacobian}) and
$\nabla_{\mathbf{v}_t}\mathbf{f}_t =
\text{diag}(\{\nabla_{\bm{v}_{t,i}}\bm{f}_{t,i}\})\in\R^{3n_c\times3n_c}$, with
each block computed as the gradient with respect to $\bm{v}_{t,i}$ in Eq.
(\ref{eq:regularized_friction_model}):
\begin{align}
    \nabla_{\bm{v}_{t,i}}\bm{f}_{t,i} = -\pi_{i,0}\left[
        \frac{\tilde{\mu}(s_i)}{\Vert\bm{v}_{t,i}\Vert}\mathbf{P}_n^\perp(\qs)+
        \frac{\tilde{\mu}'(s_i)}{v_s}\mathbf{P}_n(\qs)
    \right],
\end{align}
where $\tilde{\mu}'(s)=\mathrm d\tilde{\mu}/\mathrm ds$. We introduce the
tangential direction Delassus operator as $\mathbf{W}_{tt,0} =
-\mathbf{J}^T_{t,0}\nabla_{\mathbf{v}_t}\mathbf{f}_t(\qs_0,
\vs)\mathbf{J}_{t,0}$ and write the Jacobian of the residual as
\begin{equation}
    \nabla_\mathbf{v}\mathbf{r}(\vs) = \mathbf{M}_0 + h\,\mathbf{W}_{tt,0}.
    \label{eq:one_way_coupling_jacobian_with_delassus}
\end{equation}
We note that when $\tilde{\mu}'(s) \ge 0$, $-\nabla_{\bm{v}_{t,i}}\bm{f}_{t,i}$ is a linear
combination of positive semi-definite matrices and thus
$\mathbf{W}_{tt,0}\succeq 0$. Therefore
$\nabla_\mathbf{v}\mathbf{r}(\vs)\succ 0$, and its inverse always exists.

\subsection{Implicit Approximation to Normal Forces}
We present an approximation to treat normal forces implicitly while still
keeping the configurations \textit{frozen}. For simplicity, we consider
frictionless contact first
\begin{equation}
    \mathbf{r}(\vs) = \mathbf{M}_0\vs - \mathbf{p}^* - h\,\mathbf{J}^T_{c,0}\mathbf{f}_n(\qs, \vs),
    \label{eq:frictionless_residual}
\end{equation}
where notice we decided to \textit{freeze} the contact Jacobian but not the
compliant contact forces.

To treat the normal contact forces implicitly while still freezing the
configuration at $\qs_0$, we propose the following first-order estimation of the
penetration distance $\delta_i$ at the \nth{i} contact point,
\begin{equation}
    \delta_i \approx \delta_{i,0} - h\,v_{n,i},
    \label{eq:penetration_approximation}
\end{equation}
where we used the fact that $\dot{\delta}_i = -v_{n,i}$. We can then substitute
Eq. (\ref{eq:penetration_approximation}) into Eq.
(\ref{eq:compliant_normal_force}) to express the compliant forces,
$\pi_i(v_{n,i})$, as function of the normal velocities and group them into a
vector, $\bm{\pi}(\mathbf{v}_n)\in \R^{n_c}$. We then express the normal forces
as $\mathbf{f}_n(\mathbf{v}_n) =\hat{\mathbf{N}}_0\bm{\pi}(\mathbf{v}_n)$. This
can then be used in Eq. (\ref{eq:frictionless_residual}) to compute the Jacobian
of the residual,
\begin{align}
    \nabla_\mathbf{v}\mathbf{r}(\vs)
    &= \mathbf{M}_0 - h\,\mathbf{J}^T_{c,0}\nabla_\mathbf{v}\mathbf{f}_n(\mathbf{v}_n),\nonumber\\
    &= \mathbf{M}_0 - h\,\mathbf{J}^T_{c,0}\nabla_{\mathbf{v}_n}\mathbf{f}_n(\mathbf{v}_n)\,\mathbf{J}_{n,0},\nonumber\\
    &= \mathbf{M}_0 - h\,\mathbf{J}^T_{c,0}\hat{\mathbf{N}}_0\nabla_{\mathbf{v}_n}\bm{\pi}(\mathbf{v}_n)\,\mathbf{J}_{n,0},\nonumber\\
    &= \mathbf{M}_0 + h\,\mathbf{W}_{nn,0},
    \label{eq:frictionless_jacobian}
\end{align}
where we introduced the normal direction Delassus operator $\mathbf{W}_{nn,0} =
-\mathbf{J}^T_{n,0}\nabla_{\mathbf{v}_n}\bm{\pi}(\mathbf{v}_n)\,\mathbf{J}_{n,0}$
and $\nabla_{\mathbf{v}_n}\bm{\pi}(\mathbf{v}_n) =
\text{diag}(\{\mathrm{d}\pi_i/\mathrm{d}v_{n,i}\})$. Eq. (\ref{eq:compliant_normal_force}) implies
$\mathrm{d}\pi_i/\mathrm{d}v_{n,i} \le 0$,  and thus
$\mathbf{W}_{nn,0}\succeq 0$. Therefore in Eq.
(\ref{eq:frictionless_jacobian}) $\nabla_\mathbf{v}\mathbf{r}(\vs)\succ 0$, and its inverse always exists,
explaining the high stability of this scheme for contact problems without
friction.

\subsection{TAMSI: Two-Way Coupled Scheme}
TAMSI is a semi-implicit Euler scheme in the friction forces as
introduced in Section \ref{sec:one_way_coupling} modified to implicitly couple
the compliant contact forces using the approximation in Eq.
(\ref{eq:penetration_approximation}). It is a \textit{two-way coupled} scheme in
that, in addition to normal forces feeding into the computation of the friction
forces through Eq. (\ref{eq:regularized_friction_model}), friction forces
feedback implicitly into the normal forces.

Freezing the position kinematics to $\qs_0$ and using the approximation in Eq.
(\ref{eq:penetration_approximation}), the full TAMSI residual becomes
\begin{equation}
    \mathbf{r}(\vs) = \mathbf{M}_0\vs - \mathbf{p}^*
    -h\,\mathbf{J}^T_{n,0}\bm{\pi}(\mathbf{v}_n)
    -h\,\mathbf{J}^T_{t,0}\mathbf{f}_t(\mathbf{v}_t).
    \label{eq:tamsi_residual}
\end{equation}

We can then use the results from the previous sections, except we also need to
take into account how the friction forces in Eq.
(\ref{eq:regularized_friction_model}) change due to changes in the normal forces.
Instead of the result in Eq. \eqref{eq:jacobian_of_tangential_torque}, we now
have
\begin{align}
    \nabla_{\vs}\bm{\tau}_t &= \mathbf{J}_{t,0}^T\,\nabla_{\vs}\mathbf{f}_t,\nonumber\\
    &=\mathbf{J}_{t,0}^T\,\nabla_{\mathbf{v}_t}\mathbf{f}_t\,\mathbf{J}_{t,0} +
    \mathbf{J}_{t,0}^T\,\nabla_{\mathbf{v}_n}\mathbf{f}_t\,\mathbf{J}_{n,0},\nonumber\\
    &= -\mathbf{W}_{tt} - \mathbf{W}_{tn},
\end{align}
where $\nabla_{\mathbf{v}_n}\mathbf{f}_t =
-\text{diag}(\{\tilde{\mu}_i\hat{\bm{v}}_{t,i}\mathrm{d}\pi_i/\mathrm{d}v_{n,i}\}) \in \R^{3 n_c \times n_c}$.
We define the generally non-symmetric operator $\mathbf{W}_{tn} =
-\mathbf{J}_{t,0}^T\,\nabla_{\mathbf{v}_n}\mathbf{f}_t\,\mathbf{J}_{n,0}
\in\R^{n_v\times n_v}$ which introduces the additional two-way coupling between
compliance in the normal direction and friction in the tangential direction.

Using these results we can write the Jacobian of the TAMSI scheme as
\begin{equation}
    \nabla_\mathbf{v}\mathbf{r}(\vs) = 
    \mathbf{M}_0 + h\,\left(\mathbf{W}_{nn,0} + \mathbf{W}_{tt,0} + \mathbf{W}_{tn,0}\right).
    \label{eq:tamsi_jacobian_with_delassus}
\end{equation}

Notice that, due to $\mathbf{W}_{tn,0}$, the Jacobian is not symmetric and in
general not invertible; however, since $\mathbf{M}_0\succ 0$, the Jacobian is
invertible for sufficiently small $h$. 

After computing the next Newton iteration using this Jacobian, we use TALS to
selectively backtrack the iteration.

\subsection{Implementation Details}
We implemented a single-threaded TAMSI in the open-source robotics toolbox Drake
\cite{bib:drake}. At this moment, our initial
implementation did not focus on performance but rather on proving the stability and
robustness of the method. Therefore our initial implementation forms the
Newton-Raphson Jacobian in Eq. (\ref{eq:tamsi_jacobian_with_delassus})
explicitly. Forming $\mathbf{M}_0$ is an $\mathcal{O}(n_v^2)$ operation, while
computing the Delassus operators in Eq. (\ref{eq:tamsi_jacobian_with_delassus})
via explicitly multiplying the Jacobians is $\mathcal{O}(n_v^2\cdot n_c)$. The
factorization of $\nabla_\mathbf{v}\mathbf{r}$ is an
$\mathcal{O}(n_v^3)$ operation and we expect it to dominate for large scenarios.
However, even in our largest simulations with hundreds of DOFs and hundreds of
contacts \cite{bib:TRImanipulation}, forming this Jacobian is still the most expensive
operation, followed by its factorization. Performance can be greatly improved by
using matrix-free operators instead of explicitly forming the contact Jacobians,
allowing the assembly of $\nabla_\mathbf{v}\mathbf{r}$ in
$\mathcal{O}(n_v^2)$. Further improvements include exploiting the sparsity
pattern of $\nabla_\mathbf{v}\mathbf{r}$ during both assembly and factorization
as well as parallelization for large scenes with hundreds of objects.

\section{Results and Discussion}
\label{sec:results}

We present a series of simulation test cases to assess robustness, accuracy and
performance. For all cases we use a regularized friction parameter of
$v_s=10^{-4}~\text{m}/\text{s}$, which results in negligible sliding during
stiction. 

\subsection{Parallel Jaw Gripper}

\begin{figure}
    \centering
    \vspace{6pt}
    \includegraphics[width=0.57\columnwidth]{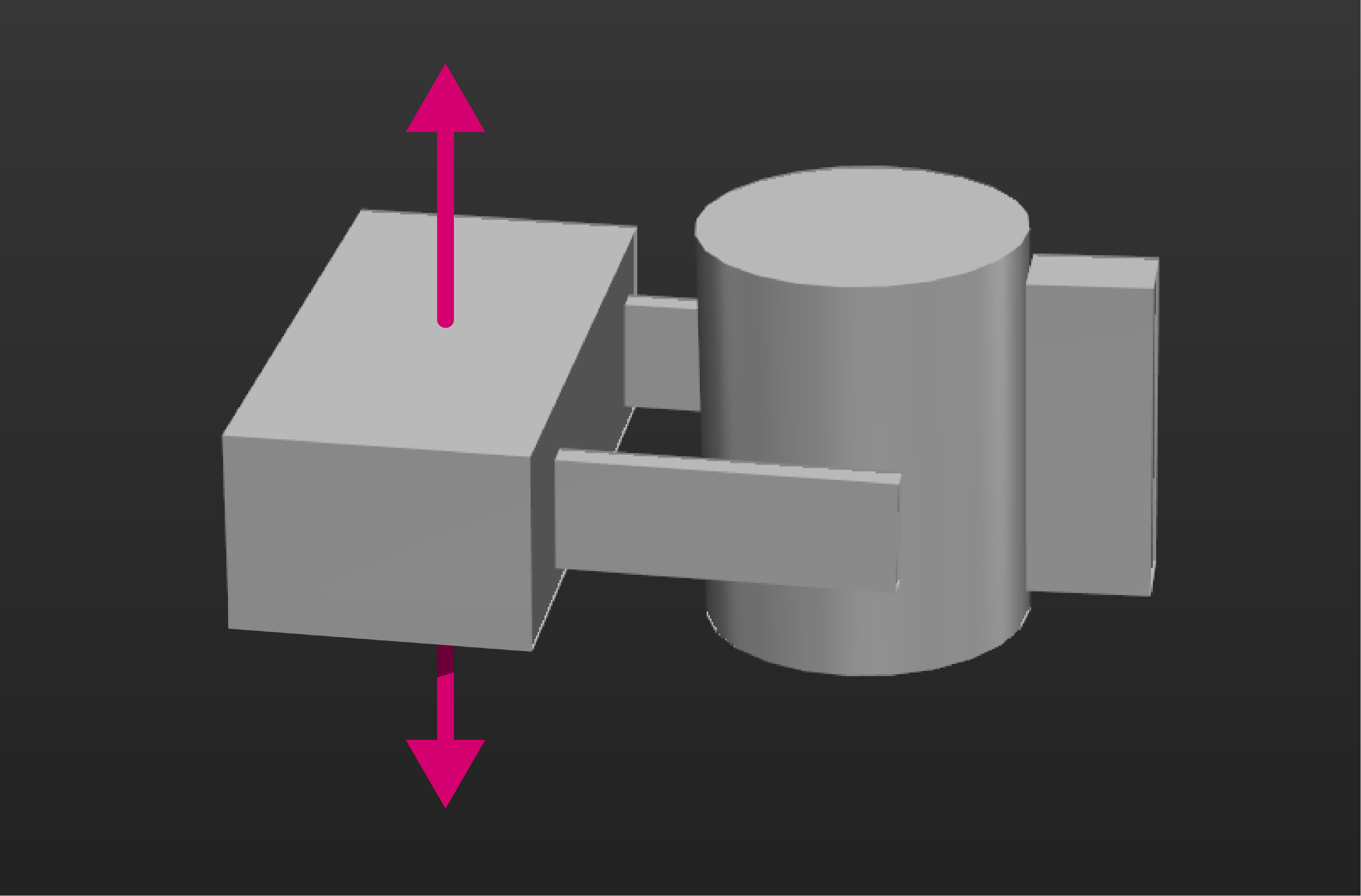}
    \includegraphics[width=0.37\columnwidth]{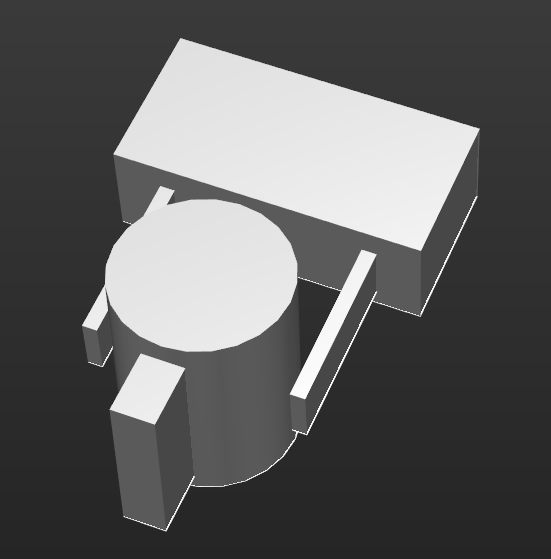}
    \caption{
        Parallel jaw gripper model. We shake the mug vertically to assess
        robustness.
    }
    \label{fig:simplegripperpicture}
\end{figure}

To assess the robustness of our method in a relevant manipulation task with large
external disturbances, we simulate a parallel jaw gripper holding a mug
and forced to oscillate vertically with a period of $T=0.5~\text{s}$ and an
amplitude of $A=15~\text{cm}$, see Fig. \ref{fig:simplegripperpicture}. To
stress the solver in a situation with slip-stick transitions, we chose a low
coefficient of friction of $\mu=0.1$ and a grip force of $10~\text{N}$. The mug
is $10~\text{cm}$ tall with a radius of $4~\text{cm}$ and a mass of
$100~\text{g}$. There is no gravity.

\begin{figure}
    \centering
    \includegraphics{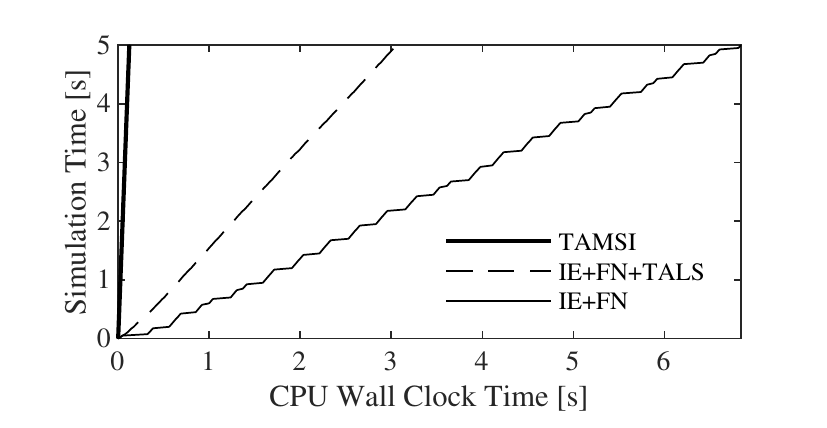}
    \caption{
        Simulated time vs. wall-clock time for the parallel jaw gripper case.
        All three runs use $h=3$~ms. TAMSI is the fastest, on the far left. The
        horizontal plateaus in the IE+FN run indicate that without TALS, the
        integrator slows down since it takes smaller step sizes for convergence
        control.}
    \label{fig:simplegripperruntimes}
\end{figure}

\begin{figure}
    \centering
    \includegraphics[scale=0.7]{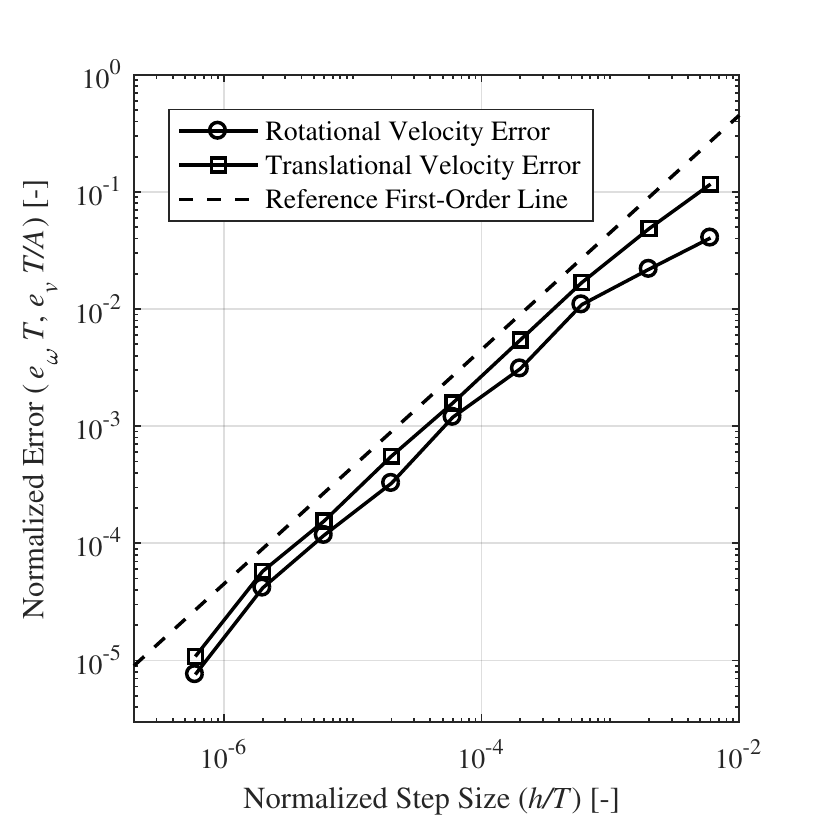}
    \caption{
        $L^2\text{-norm}$ of the generalized velocity errors
        against time step, nondimensionalized with the problem's period $T$ and
        amplitude $A$. Both translational and rotational DOFs exhibit
        first-order convergence as expected.}
    \label{fig:simplegrippertamsiconvergence}
\end{figure}

With a step $h=3~\text{ms}$ TAMSI completes 5~seconds of simulation time with a
single thread on an Intel i7-6900K 3.2 GHz CPU in 123~ms of wall-clock time, or
40$\times$ real-time rate. In contrast, IE+FN with TALS is $25\times$ slower
than TAMSI and IE+FN without TALS is $55\times$ slower than TAMSI. Simulated time vs
wall-clock time for each is shown in Fig. \ref{fig:simplegripperruntimes}. Next,
we perform a convergence study of TAMSI running with different
time steps and estimate errors against a reference solution using
$h=10^{-7}~\text{s}$, Fig. \ref{fig:simplegrippertamsiconvergence}. As expected,
TAMSI exhibits first-order convergence with step size.

\subsection{Allegro Hand}
\begin{figure}
    \centering
    \includegraphics[width=0.47\columnwidth]{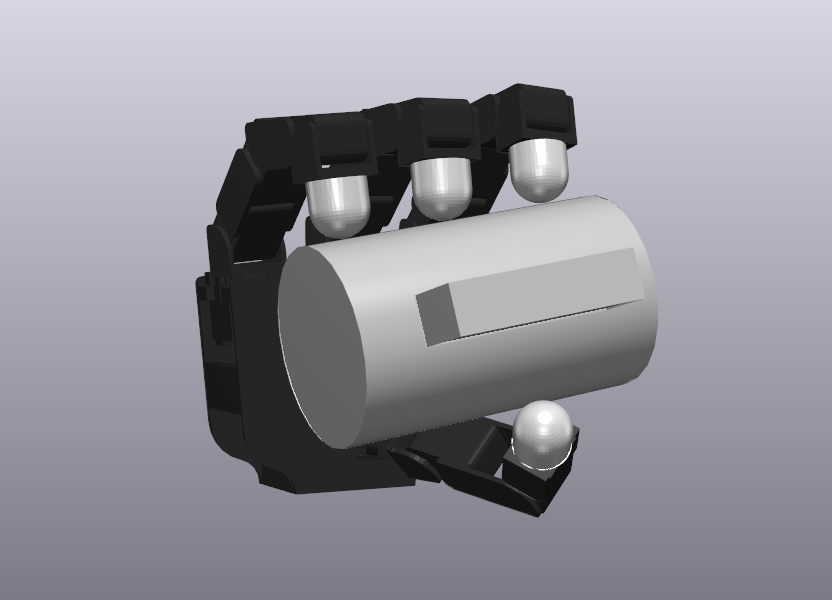}
    \includegraphics[width=0.51\columnwidth]{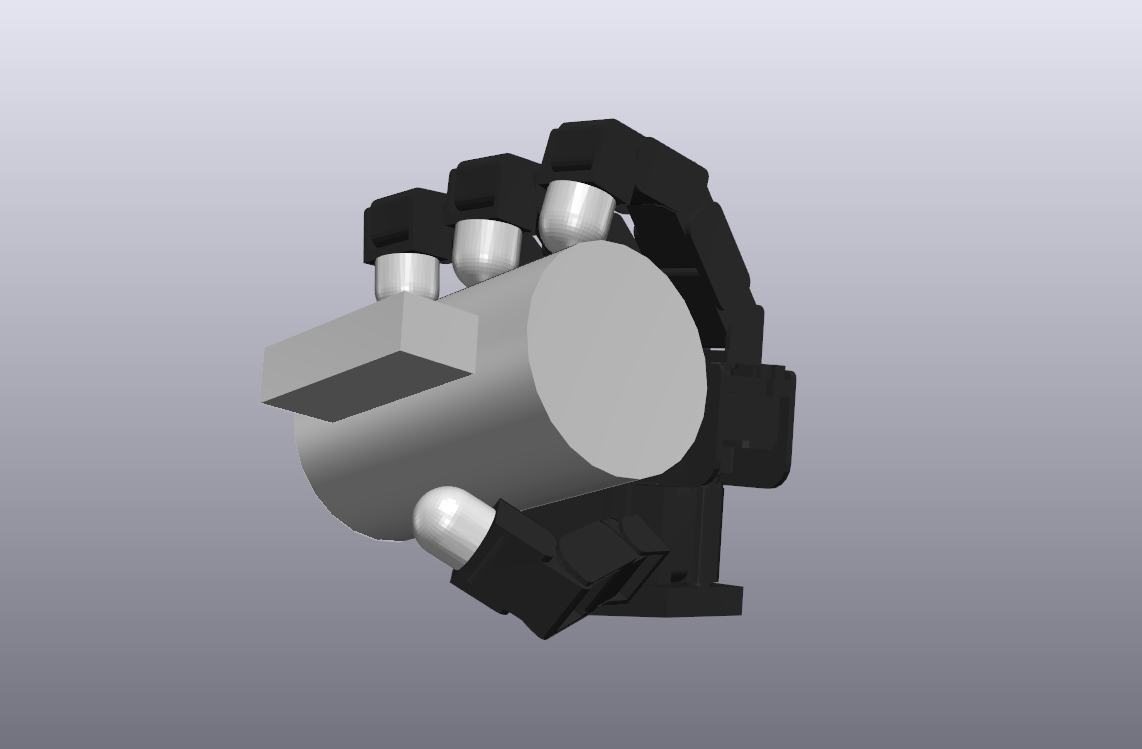}
    \caption{
        Model of a 16 DOF Allegro hand holding a mug. As the hand reorients the
        mug, multiple simultaneous points of contact securing the grasp are
        made.
        }
    \label{fig:allegrohandpicture}
\end{figure}

We simulate a 16 DOF Allegro hand controlled in open-loop to
perform a periodic reorientation of a mug, see Fig. \ref{fig:allegrohandpicture}
and the supplemental video accompanying this manuscript. This interesting system
includes multiple points of contact, complex geometry and a large number of
DOFs. 

We observe that TALS does not improve the performance of implicit Euler --- as
the system configuration $\mathbf{q}$ changes, the tangent space reorients, and
TALS is unable to properly control iterations with transitions into stiction.
TAMSI however, with a time step $h=0.7~\text{ms}$, runs $7.8\times$ faster than
our fastest integrator setup, the error-controlled implicit Euler using
quasi-Newton when solving to a (loose) accuracy of $a=0.01~\text{m}/\text{s}$. Our single-threaded TAMSI
completes 15 seconds simulated time on an Intel i7-6900K 3.2 GHz CPU in 7.0~s of
wall-clock time, or $2.14\times$ real-time rate.

\subsection{Sim-to-Real Experiments}
\label{sec:sim_to_real}

We demonstrate our simulation capability in two manipulation tasks
with closed-loop control involving force-feedback and manipuland-world contact.

A Kuka IIWA arm (7 DOF) is outfitted with a Schunk WSG 50 gripper. We use an
inverse dynamics controller with gains in the simulation set to best match
reality, even though the specifics of Kuka's proprietary joint-level controller
are unavailable. The controller process tracks a prescribed sequence of
Cartesian end-effector keyframe poses and computes desired joint-space
trajectories using Jacobian IK. We use force feedback to regulate the grasp
force and judge for placement success. 

The gripper surfaces are slightly compliant. We identify the stiffness
parameters of the point contact model by matching the level of compliance
observed in the real hardware.

In the first task the robot is commanded to grab a
water bottle, perform a vigorous shaking motion, and place the
bottle at a new location, demonstrating the robustness of the method to strong
disturbances; see the supplemental video.

In the second task a Kuka arm in a kitchen scenario grabs a mug and uses
external contact with the sink to help reorient the mug, relying on slip
within the fingers of the gripper; see Fig. \ref{fig:teaser} and the
accompanying supplemental video. TAMSI robustly handles multiple points of
contact, changes in the contact configuration, and stick-slip
transitions.

TAMSI enabled us to prototype this task entirely in simulation with controllers
that transitioned seamlessly to reality.

\section{Conclusions}
\label{sec:conclusion}

In this work we systematically analyzed implicit integration for multibody
problems with compliant contact and regularized friction. Error-controlled
implicit Euler with quasi-Newton iterations performs best, though it spends a
significant amount of time resolving the high-frequency transition
dynamics introduced by regularized friction. Our new TALS method helps to
address this problem, though its performance degrades as the configuration of
the system changes and contact surfaces reorient. This observation led us to
develop our novel TAMSI method. TAMSI approximates penetration distances to
first-order to couple contact forces implicitly while
performing only a single geometric query at the beginning of a time step, resulting
in improved performance. We demonstrate the added robustness and performance of
TAMSI with simulations of relevant manipulation tasks. Sim-to-real comparisons
show the usage of TAMSI to prototype controllers in simulation that
transfer effectively to reality.

Additional examples can be found in the ``examples'' source directory
of the open-source robotics toolbox Drake
\cite{bib:drake} and in the ``Drake Gallery'' section of the documentation.

Ongoing work conducted at the Toyota Research Institute is
leveraging the proposed method for prototyping and validating controllers for
robot manipulation in dense cluttered environments \cite{bib:TRImanipulation}
and extending TAMSI to work with more sophisticated contact models
\cite{bib:elandt2019pressure}.

\section*{Acknowledgment}
We thank the reviewers for their feedback, Allison Henry and 
Siyuan Feng for technical support, and Evan
Drumwright and Russ Tedrake for helpful discussions. The work of Ante Qu was supported in part by
the National Science Foundation under grant DGE-1656518. Any opinions, findings,
and conclusions or recommendations expressed in this material are those of
the authors and do not necessarily reflect the views of the National Science Foundation.

\bibliographystyle{IEEEtran}
\bibliography{transition_aware_method}


\addtolength{\textheight}{-12cm}   



\end{document}